\documentclass{article} 
\usepackage{nips15submit_e,times}
\usepackage{hyperref}
\usepackage{url}
\usepackage{cite}
\usepackage{graphicx}

\title{Generating News Headlines with Recurrent Neural Networks}

\author{Konstantin Lopyrev \\
klopyrev@stanford.edu}

\nipsfinalcopy 

\begin{document}

\maketitle

\begin{abstract}
We describe an application of an encoder-decoder recurrent neural network with LSTM units and attention to generating headlines from the text of news articles.
We find that the model is quite effective at concisely paraphrasing news articles.
Furthermore, we study how the neural network decides which input words to pay attention to, and specifically we identify the function of the different neurons in a simplified attention mechanism.
Interestingly, our simplified attention mechanism performs better that the more complex attention mechanism on a held out set of articles.

\end{abstract}

\section{Background}
Recurrent neural networks have recently been found to be very effective for many transduction tasks - that is transforming text from one form to another.
Examples of such applications include machine translation\cite{DBLP:journals/corr/SutskeverVL14, DBLP:journals/corr/LuongPM15} and speech recognition \cite{DBLP:journals/corr/abs-1303-5778}.
These models are trained on large amounts of input and expected output sequences, and are then able to generate output sequences given inputs never before presented to the model during training.

Recurrent neural networks have also been applied recently to reading comprehension \cite{DBLP:journals/corr/HermannKGEKSB15}.
There, the models are trained to recall facts or statements from input text.

Our work is closely related to \cite{DBLP:journals/corr/RushCW15} who also use a neural network to generate news headlines using the same dataset as this work.
The main difference to this work is that they do not use a recurrent neural network for encoding, instead using a simpler attention-based model.

\section{Model}
\subsection{Overview}
\begin{figure}[h]
\begin{center}
\includegraphics[scale=0.18]{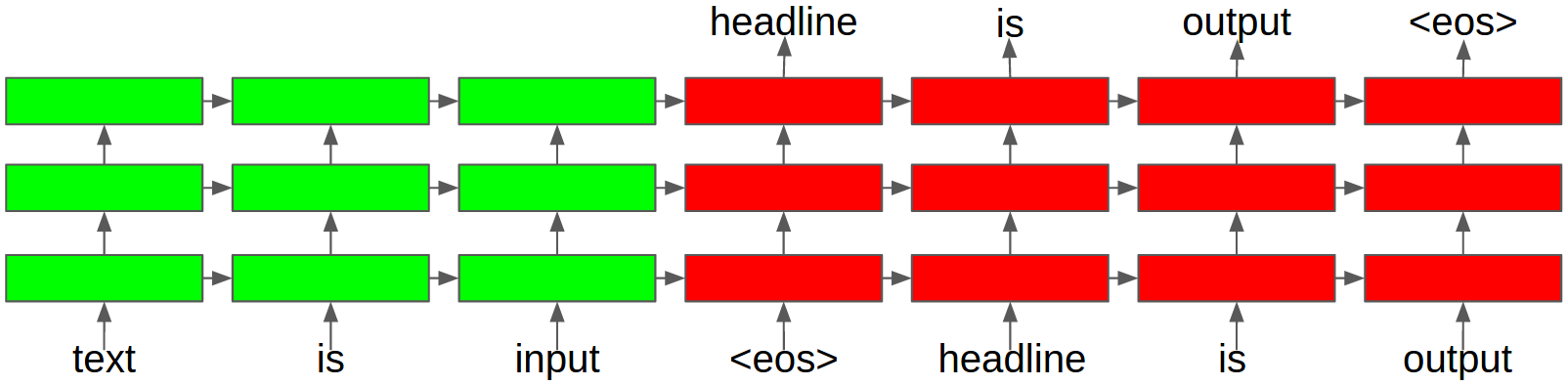}
\end{center}
\caption{Encoder-decoder neural network architecture}
\label{seqtoseq}
\end{figure}

We use the encoder-decoder architecture described in \cite{DBLP:journals/corr/SutskeverVL14} and \cite{ DBLP:journals/corr/LuongPM15}, and shown in figure \ref{seqtoseq}. The architecture consists of two parts - an encoder and a decoder - both by themselves recurrent neural networks.

The encoder is fed as input the text of a news article one word of a time.
Each word is first passed through an embedding layer that transforms the word into a distributed representation.
That distributed representation is then combined using a multi-layer neural network with the hidden layers generated after feeding in the previous word, or all 0's for the first word in the text.

The decoder takes as input the hidden layers generated after feeding in the last word of the input text.
First, an end-of-sequence symbol is fed in as input, again using an embedding layer to transform the symbol into a distributed representation.
Then, the decoder generates, using a softmax layer and the attention mechanism, described in the next section, each of the words of the headline, ending with an end-of-sequence symbol.
After generating each word that same word is fed in as input when generating the next word.

The loss function we use is the log loss function:
\begin{eqnarray*}
-\log p(y_1, \ldots, y_{T'} | x_1, \ldots, x_T) = - \sum_{t = 1}^{T'} \log p(y_t | y_1, \ldots, y_{t - 1}, x_1, \ldots, x_T)
\end{eqnarray*}
where $y$ represent output words and $x$ represent input words.

Note that during training of the model it is necessary to use what is called ``teacher forcing'' \cite{Goodfellow-et-al-2015-Book}.
Instead of generating a new word and then feeding in that word as input when generating the next word, the expected word in the actual headline is fed in.
However, during testing the previously generated word is fed in when generating the next word.
That leads to a disconnect between training and testing.
To overcome this disconnect, during training we randomly feed in a generated word, instead of the expected word, as suggested in \cite{DBLP:journals/corr/BengioVJS15}.
Specifically, we do this 10\% of the time, as also done in \cite{DBLP:journals/corr/ChanJLV15}.
During testing we use a beam-search decoder which generates input words one at a time, at each step extending the $B$ highest probability sequences.

We use 4 hidden layers of LSTM units, specifically the variant described in \cite{DBLP:journals/corr/ZarembaSV14}.
Each layer has 600 hidden units.
We attempted using dropout as is also described in \cite{DBLP:journals/corr/ZarembaSV14}.
However we did not find it to be useful.
Thus, the models analyzed below do not use dropout.
We initialize most parameters of the model uniformly in the range $[-0.1; 0.1]$.
We initialize the biases for each word in the softmax layer to the log-probability of its occurence in the training data, as suggested in \cite{DBLP:journals/corr/KarpathyF14}.

We use a learning rate of 0.01 along with the RMSProp \cite{RMSPROP} adaptive gradient method.
For RMSProp we use a decay of 0.9 and a momentum of 0.9.
We train for 9 epochs, starting to half the learning rate at the end of each epoch after 5 epochs.

Additionally, we batch examples, processing 384 examples at a time.
This batching complicates the implementation due to the varying lengths of different sequences.
We simply fix the maximum lengths of input and output sequences and use special logic to ensure that the correct hidden states are fed in during the first step of the decoder, and that no loss is incurred past the end of the output sequence.

\subsection{Attention}
Attention is a mechanism that helps the network remember certain aspects of the input better, including names and numbers.
The attention mechanism is used when outputting each word in the decoder.
For each output word the attention mechanism computes a weight over each of the input words that determines how much attention should be paid to that input word.
The weights sum up to 1, and are used to compute a weighted average of the last hidden layers generated after processing each of the input words.
This weighted average, referred to as the context, is then input into the softmax layer along with the last hidden layer from the current step of the decoding.

\begin{figure}[h]
\centering
\begin{minipage}{.5\textwidth}
\centering
\includegraphics[scale=0.2]{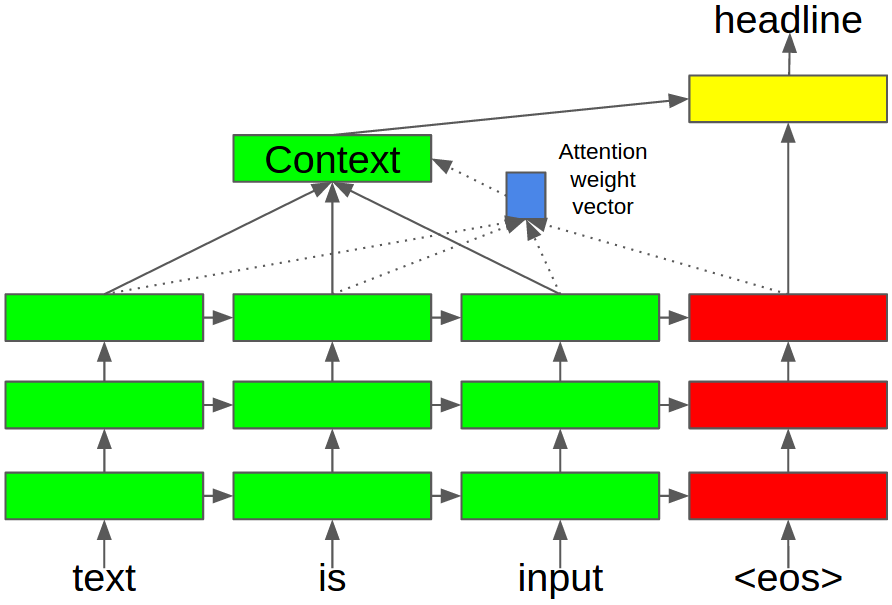}
\caption{Complex attention}
\label{complex-attention}
\end{minipage}%
\begin{minipage}{.5\textwidth}
\centering
\includegraphics[scale=0.2]{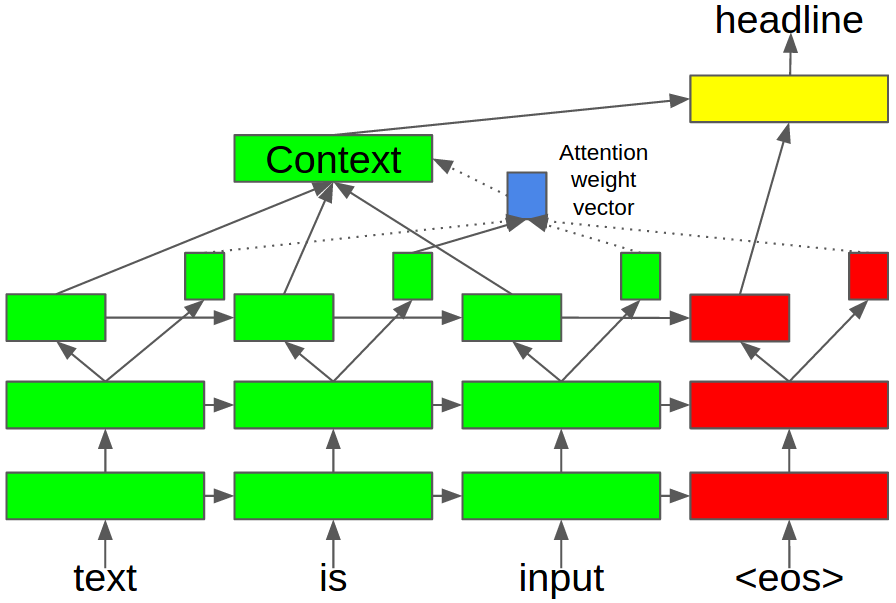}
\caption{Simple attention}
\label{simple-attention}
\end{minipage}
\end{figure}

We experiment with two different attention mechanisms.
The first attention mechanism, which we refer to as \textit{complex} attention, is the same as the \textit{dot} mechanism in \cite{DBLP:journals/corr/LuongPM15}.
This mechanism is shown in figure \ref{complex-attention}.
The attention weight for the input word at position $t$, computed when outputting the $t'$-th word is:
\begin{eqnarray*}
	a_{y_{t'}}(t) = \frac{\exp(h_{x_{t}}^T h_{y_{t'}})}{\sum_{\bar t}^T \exp(h_{x_{\bar t}}^T h_{y_{t'}})}
\end{eqnarray*}
where $h_{x_{t}}$ represents the last hidden layer generated after processing the $t$-th input word, and $h_{y_{t'}}$ represents the last hidden layer from the current step of decoding.
Note one of the characteristics of this mechanism is that the same hidden units are used for computing the attention weight as for computing the context.

The second attention mechanism, which we refer to as \textit{simple} attention, is a slight variation of the \textit{complex} mechanism that makes it easier to analyze how the neural network learns to compute the attention weights.
This mechanism is shown in figure \ref{simple-attention}.
Here, the hidden units of the last layer generated after processing each of the input words are split into 2 sets: one set of size 50 used for computing the attention weight, and the other of size 550 used for computing the context.
Analogously, the hidden units of the last layer from the current step of decoding are split into 2 sets: one set of size 50 used for computing the attention weight, and the other of size 550 fed into the softmax layer.
Aside from these changes the formula for computing the attention weights, given the correponding hidden units, and the formula for computing the context are kept the same.

\section{Dataset}
\subsection{Overview}
The model is trained using the English Gigaword dataset, as available from the Stanford Linguistics department.
This dataset consists of several years of news articles from 6 major news agencies, including the New York Times and the Associated Press.
Each of the news articles has a clearly delineated headline and text, where the text is broken up into paragraphs.
After the preprocessing described below the training data consists of 5.5M news articles with 236M words.
\subsection{Preprocessing}
The headline and text are lowercased and tokenized, separating punctuation from words.
Only the first paragraph of the text is kept.
An end-of-sequence token is added to both the headline and the text.
Articles that have no headline or text, or where the headline or text lengths exceed 25 and 50 tokens, respectively, are filtered out, for computational efficiency purposes.
All rare words are replaced with the $<$ unk $>$ symbol, keeping only the 40,000 most frequently occuring words.

The data is split into a training and a holdout set.
The holdout set consists of articles from the last month of data, with the second last month not included in either the training or holdout sets.
This split helps ensure that no nearly duplicate articles make it into both the training and holdout sets.

Finally, the training data is randomly shuffled.

\subsection{Dataset Issues}
The dataset as used has a number of issues.
There are many training examples where the headline does not in fact summarize the text very well or at all.
These include many articles that are formatted incorrectly, having the actual headline in the text section and the headline section containing words such as ``(For use by New York Times News service clients)''.
There are many articles where the headline has some coded form, such as ``biz-cover-1stld-writethru-nyt`` or ``bc-iraq-post 1stld-sub-pickup4thgraf``.

No filtering of such articles was done.
An ideal model should be able to handle such issues automatically, and attempts were made to do so using, for example, randomly feeding in generated words during training, as described in the Model section.

\section{Evaluation}
The performance of the model was measured in two different ways.
First, we looked at the training and holdout loss.
Second, we used the BLEU \cite{Papineni:2002:BMA:1073083.1073135} evaluation metric over the holdout set, defined next.
For efficiency reasons, the holdout metrics were computed over only 384 examples.

The BLEU evaluation metric looks at what fraction of n-grams of different lengths from the expected headlines are actually output by the model.
It also considers the number of words generated in comparison to the number of words used in the expected headlines.
Both of these are computed over all 384 heldout example, instead of over each example separately.
For the exact definition see  \cite{Papineni:2002:BMA:1073083.1073135}.

\section{Analysis}
\begin{figure}[h]
\centering
\begin{minipage}{.5\textwidth}
\centering
\includegraphics[scale=0.4]{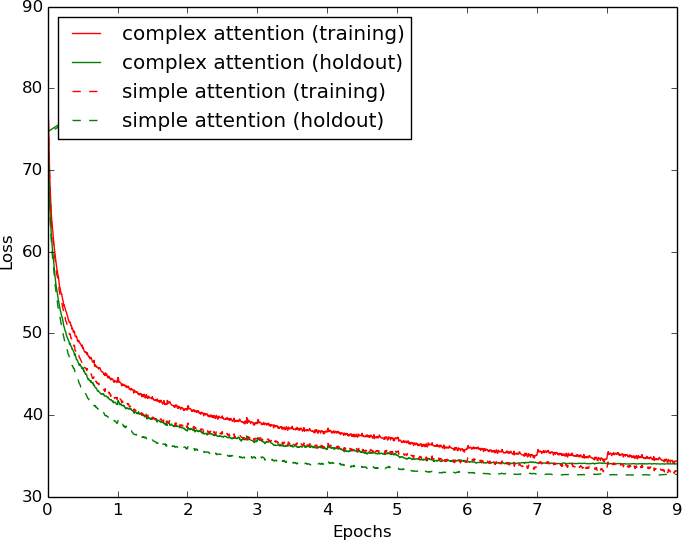}
\caption{Loss vs epoch}
\label{loss}
\end{minipage}%
\begin{minipage}{.5\textwidth}
\centering
\includegraphics[scale=0.4]{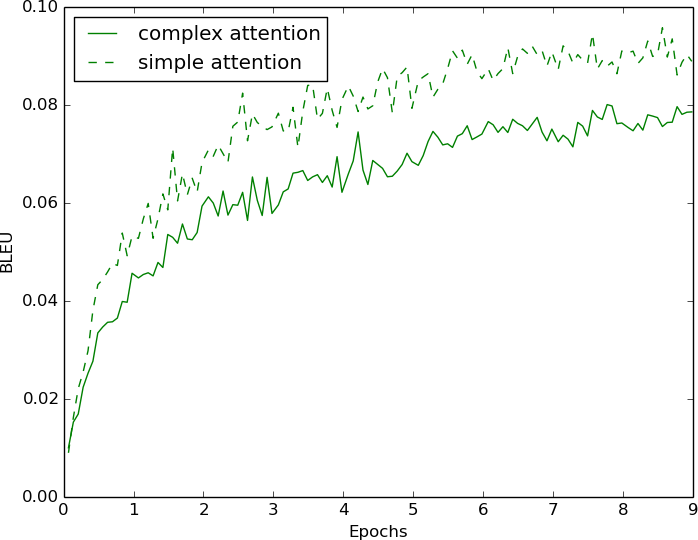}
\caption{BLEU vs epoch}
\label{bleu}
\end{minipage}
\end{figure}

Each model takes 4.5 days to train on a GTX 980 Ti GPU.
Figures \ref{loss} and \ref{bleu} show the evaluation metrics as a function of training epoch.
Note that in our setup the training loss is generally higher than holdout loss, since when computing the holdout loss we don't feed in generated words 10\% of time.

The model is quite effective in predicting headlines from the same newspapers as it was trained on.
Table \ref{example-predictions} lists 7 examples chosen at random from the held-out examples.
The model generally seems to capture the gist of the text and manages to paraphrase the text, sometimes using completely new words.
However, it does make mistakes, for example, in sentences 2, 4 and 7.

\begin{table}[t]
\caption{Example predictions}
\label{example-predictions}
\begin{center}
\begin{tabular}{|p{2.6in}  p{1.2in}  p{1.2in}|}
\hline
\textbf{Text} & \textbf{Actual Headline} & \textbf{Predicted Headline} \\
\hline
1. At least 72 people died and scores more were hurt when a truck crowded with pilgrims plunged into a gorge in the desert state of Rajasthan on Friday, police told the press trust of India.
&
Urgent: truck crashes killing 72 pilgrims in India
&
At least 72 dead in Indian road accident
\\
 \hline
2. Sudanese president Omer Al-Bashir has announced his refusal of discharging a government minister who had been accused by the International Criminal Court (ICC) of committing war crimes in the Western Sudanese region of Darfur, Sudan's $<$unk$>$ Daily reported on Monday.
&
Sudanese president refuses to discharge state minister indicted by ICC
&
Sudanese president refuses to of alleged war crimes
 \\
\hline
3. A chief of Afghanistan's ousted Taliban militia said Al-Qaeda chief Osama Bin Laden is alive and has sent him a letter of condolences, in an interview broadcast on Tuesday on Al-Jazeera television.
&
Taliban leader says Bin Laden still alive
&
Urgent: Bin Laden alive, says Taliban chief
\\
\hline
4. One of the last remaining routes for Iraqis trying to flee their country has effectively been closed off by new visa restrictions imposed by Syria, the U. N. refugee agency said Tuesday.
&
UNHCR says new Syrian visa rules blocking Iraqis from entering country 
&
U.N. refugee agency closes last routes to Iraq
\\
\hline
5. Members of the U.N.'s new human rights watchdog on Tuesday formally adopted a series of reforms to its future work, including how and when to launch investigations into some of the world's worst rights offenders.
&
U.N. human rights watchdog adopts reforms on how to investigate countries for abuses
&
U.N. human rights body adopts reforms 
\\
\hline
6. Democratic presidential candidates said Thursday they would step up pressure on Pakistan's president Pervez Musharraf over democracy, and criticized White House policy towards Islamabad.
&
Democrats call for more pressure on Pakistan 
&
Democratic presidential hopefuls call for pressure on Musharraf
\\
\hline
7. Manchester United's strength in depth is set to be tested for the first time this season in the wake of a last-gasp win over Liverpool that has significantly shortened the odds on Sir Alex Ferguson's side reclaiming the premiership title.
&
Football: United face test of reserves after Scholes red card by Neil Johnston
&
United set for test test
\\
\hline
\end{tabular}
\end{center}
\end{table}

The model has much more mixed performance when used to generate headlines for news articles from sources that are different from training.
Table \ref{example-predictions-new-sources} shows generated headlines for articles from several major news websites.
The model does quite well with articles from the BBC, the Wall Street Journal and the Guardian.
However, it performs very poorly on articles from the Huffington Post and Forbes.
In fact, the model performed poorly on almost all tested articles from Forbes.
It seems that there is a major difference in how articles from Forbes are written, when compared to articles used to train the model.

\begin{table}[t]
\caption{Example predictions from sources different from training}
\label{example-predictions-new-sources}
\begin{center}
\begin{tabular}{|p{0.45in} p{2.1in}  p{1.13in}  p{1.13in}|}
\hline
\textbf{Source } & \textbf{Text} & \textbf{Actual Headline} & \textbf{Predicted Headline} \\
\hline
BBC
&
Russia's President Vladimir Putin has condemned Turkey's shooting down of a Russian warplane on its border with Syria.
&
Turkey's downing of Russian warplane - what we know
&
Putin condemns Turkey's shooting of Russian plane
\\
\hline
Huffington Post
&
When Sarah Palin defended governors who are refusing to accept refugees by claiming there's no vetting process to keep out terrorists, "Late Night" host Seth Meyers completely shut down her argument.
&
Seth Meyers Calls Out Sarah Palin For Repeating Refugee Lies
&
'Night of the night' is $<$unk$>$
\\
\hline
Wall Street Journal
&
The top commander of U.S. troops in Afghanistan said Wednesday that the American service members most closely associated with the deadly bombing of a Doctors Without Borders hospital in Afghanistan have been suspended from duty.
&
U.S. Troops Suspended After Afghan Hospital Bombing
&
U.S. commander in Afghanistan suspended from hospital
\\
\hline
Forbes
&
Tuesday and Wednesday, you are likely to read online or in a newspaper – and are even more likely to hear some TV or radio news person say – that the day before Thanksgiving is the busiest air travel day of the year. It long has been a staple of reporting in what usually is a very slow news week.
&
How Crazy Will Travel Be On The Day Before Thanksgiving? Not As Crazy As You've Been Led To Believe
&
The on the air
\\
 \hline
The Guardian
&
Presidential candidate Hillary Clinton has plunged into the heated debate surrounding the police killing of a black teenager in Chicago, saying “we cannot go on like this”, following the release of a video showing Laquan McDonald being shot multiple times by an officer on the street.
&
Hillary Clinton on Laquan McDonald shooting: 'We cannot go on like this' 
&
Hillary Clinton plunges into debate
\\
\hline
\end{tabular}
\end{center}
\end{table}

\subsection{Understanding information stored in last layer of the neural network}

We notice that there are multiple ways to go about understanding the function of the attention mechanism.
Consider the formula for computing the input to the softmax function:
\begin{eqnarray*}
o_{y_{t'}} = W_{co} c_{y_{t'}} + W_{ho} h_{y_{t'}} + b_o
\end{eqnarray*}
where $c_{y_{t'}}$ is the context computed for the current step of decoding, $h_{y_{t'}}$ is the last hidden layer from the current step of decoding, and $W_{co}$, $W_{ho}$ and $b_o$ are model parameters.
First, note that by looking at the word with the highest values for
\begin{eqnarray*}
W_{ho} h_{y_{t'}} + b_o
\end{eqnarray*}
we can get an idea of what exactly the hidden layer from the current step of decoding is contributing to the final generated output.
Analogously, by looking at the words with the highest values for
\begin{eqnarray*}
W_{co} c_{y_{t'}} + b_o
\end{eqnarray*}
we can do the same for the attention context.
Moreover, since the context is just a weighted sum over the hidden layers of the decoder we can compute
\begin{eqnarray*}
W_{co}  h_{x_{t}} + b_o
\end{eqnarray*}
for each of the input positions and get a good idea of what words would be recalled if the network paid attention to each of the input positions.

As an example, the last hidden layer generated after processing the ``and'' in the first example in table \ref{example-predictions} is closest to the following words:
72, at, death, in, died, dead, to, $<$eos$>$, toll, people.
We see many of the words that appeared before the ``and''.
It is interesting note that the words closest to the hidden layer for ``died'' are: 72, at, to, in, $<$eos$>$, $<$unk$>$, for, as, ``,'', more.
Note that the word ``died'' doesn't appear.
This example demonstrates our observation that the network sometimes takes multiple steps to encode a particular word.
That makes it slightly trickier to understand what the network is paying attention to.

As an example of what the hidden layer during decoding contains, the hidden layer during the first step of decoding for the same example as above is closest to the following words: urgent, (, at, one, two, three, $<$unk$>$, four, 1st, 1.
After generating the ``at'' the hidden layer during the next step is closest to: least, :, -, killed, ``,'', ld, lead, in, $<$unk$>$, to.

\subsection{Understanding how the attention weight vector is computed}
We had an initial hypothesis for how the network learned to compute the attention weight vector.
We hypothesized that the network remembered roughly which word should be generated next, and the dot product $h_{x_{t}}^T h_{y_{t'}}$ would compute the similarity between the remembered word and the actual word at the position.
For example, if the network remembered that the text talked about some number and was able to output a representation that was close to numbers in general, the attention mechanism would then allow it to get the exact number.

One implication of this hypothesis is that the units of the last hidden layer of the decoder would be in the same space as the units of the last hidden layer of the encoder. Thus, computing
\begin{eqnarray*}
W_{co}  h_{y_{t'}} + b_o
\end{eqnarray*}
and looking at the words with the highest values would tell us something meaningful.
It turned out that this was not the case.

Further analysis into which units contributed to the attention weight vector computation revealed the existence of a few units which played key roles.
That led us to simplify the neural network architecture as described in the Model section.
Interestingly, as shown in figures \ref{loss} and \ref{bleu}, the simplified model performs significantly better.
One possible explanation is that by separating the attention weight vector computation from the context computation we reduce the noise in both of these computations.

The simplified attention mechanism is also much easier to understand since only a small number of hidden units is used to compute the attention weight vector.
In one of the smaller model that we trained we used only 20 units for the attention weight vector computation, and were able to figure out some of the functions of these 20 units.
Table \ref{attention-neuron-purposes} catalogs the functions of the 20 neurons in the encoding part of the network.
Each position in the input text has 20 neurons that serve these functions.

\begin{table}[t]
\caption{Attention neuron purposes}
\label{attention-neuron-purposes}
\begin{center}
\begin{tabular}{|c|p{5in}|}
\hline
\textbf{Neuron} & \textbf{Discovered Purposes} \\
\hline
1  & Person names; country names \\
2  & Multi-part numbers (e.g. ``\$ 1 , 000''); noun after a number (e.g. ``people'' in ``four people killed''); noun after an adjective \\
3  & End of multi-part sequence (e.g. ``000'' in ``\$ 1 , 000'', ``qaida'' in ``al - qaida'', ``france'' in ``tour de france'') \\
4  & Verb after auxiliary verb or particle (e.g. ``meet'' in ``will meet'' or ``to meet''); past tense verbs; noun following a preposition  \\
5  & Beginning of text (e.g. first of two sentences) \\
6  & Verbs; prepositions \\
7  & End of noun phrase (e.g. ``finance minister'' in ``former finance minister'' or ``probe'' in ``ethics probe''); past tense verbs \\
8  & Present participles; noun phrases after ``for'' \\
9 & Objects and subjects of a verb (positive activation for object, negative for subject) \\
10  & Subjects; objects after some verbs; words after a dash \\
11 & Number following a conjuction (e.g. ``four'' in ``three people killed and four injured''); verb after auxiliary verb; noun following a preposition; \\
12  & Objects and subjects of a verb (positive activation for object, negative for subject) \\
13  & Most nouns and verbs \\
14  & Verbs; Locations; word after ``by''  \\
15  & Days of the week; some adjectives\\
16  & Function words; negations (e.g. ``not'') \\
17  & End of noun phrase \\
18  & Objects and subjects of a verb (positive activation for object, negative for subject) \\
19  & Names; some verbs; some adjectives \\
20  & Objects, subjects and corresponding verbs \\
\hline
\end{tabular}
\end{center}
\end{table}

The neural network learns to spot many linguistic phenomena.
Our analysis was also somewhat shallow due to time constraints.
We suspect that some of the neurons activate for even more complex phenomena  than we are able to find, such as different types of sentence structure.

It is important to note that the neurons in the encoding part of the network interact with the neurons in the decoding part of the network.
Indeed, the neurons in the decoding part of the network activate at different times to test for different phenomena.
For example, unit 12 starts off being positive so that the network pays attention to the object first.
Then, unit 12 becomes negative so that the network pays attention to the subject.
Interestingly, unit 9, which appears to work almost the same at unit 12 in the encoder, is usually 0 at the beginning of decoding, and only later on becomes activated.

\subsection{Errors}
The network makes a lot of different types of errors.
While we didn't do an in-depth error analysis, a few error still stood out.

One flaw with the neural network mechanism is its tendency to fill in details when details are missing.
For example, after simplifying the text given in table \ref{example-predictions}, example 1 to ``72 people died when a truck plunged into a gorge on Friday.'' the model predicts ``72 killed in truck accident in Russia''.
The model makes up the fact that the accident happened in Russia.
These errors happen most often when the number of decoding beams is small, since the model stops considering the decoding where the sentence ends early before outputing the made up details.

What also happens occasionally is the network outputs some headline that is completely unrelated to the input text
(e.g. ``urgent'', ``bc-times'' or even ``can make individual purchases by calling 212 - 556 - 4204 or - 1927 .)'').
This problem is caused by the fact that such headlines occur somewhat often in the input dataset.
These errors happen most often when the number of decoding beams is large, since that increases the probability of the model starting to generate one such high probability sequence.

These 2 examples demonstrate that our network is very sensitive to the number of decoding beams used.
We used only 2 decoding beams for the BLEU evaluation.
We suspect that if we fixed the second problem we could get much better results with a large number of beams.
One solution worthy of investigation is to use the scheduled sampling mechanism described in \cite{DBLP:journals/corr/BengioVJS15}.

\section{Future Work}
We demostrate above that the recurrent neural network learns to model complex linguistic phenomena, given large amounts of training data.
However, such large amounts of training data are usually not available for real-world NLP problems.
One interesting direction to pursue is using a dataset, like Gigaword, to pretrain a recurrent neural network that is then fine-tuned to solve a task such as part-of-speech tagging on a much smaller dataset.

More immediately, in addition to the already discussed idea to use scheduled sampling, another way to improve the model is to use a bi-directional RNN.
We suspect that the attention mechanism would work better with a bi-directional RNN, since more information would be available to model some of the phenomena outlined in table \ref{attention-neuron-purposes}.
In our current model, the network must make a decision about which values to assign the neurons used to compute the attention weight for the current input word given only the current and previous words and not any of the following words.
Giving the model information about the following words would make this decision easier for the network to make.

\section{Conclusion}
We've trained an encoder-decoder recurrent neural network with LSTM units and attention for generating news headlines using the texts of news articles from the Gigaword dataset.
Using only the first 50 words of a news article, the model generates a concise summary of those 50 words, and most of the time the summary is valid and grammatically correct.
The model doesn't perform quite as well on general text, demonstrating that a lot of the articles in Gigaword follow a particular form.

We study 2 different versions of the attention mechanism with the goal of understanding how the model decides which words of the input text to pay attention to when generating each output word.
We introduce a simplified attention mechanism that uses a small set of neurons for computing the attention weights.
This simplified mechanism makes it easier to study the function of the network.
We find that the network learns to detect linguistic phenomena, such as verbs, objects and subjects of a verb, ends of noun phrases, names, prepositions, negations, and so on.
Interestingly, we find that our simplified attention mechanism does better on the evaluation metrics.

\section*{Acknowledgements}
We'd like to thank Thang Luong for his helpful suggestions and for sharing his machine translation code.
We'd like to thank Samy Bengio and Andrej Karpathy for suggesting a few interesting design alternatives.
Finally, we'd like to thank Chris R\'e.
Some of the code written as part of doing research in his lab was used to complete this work.

\small{
\bibliography{writeup}{}
\bibliographystyle{unsrt}
}

\end{document}